\documentclass[conference]{IEEEtran}
\IEEEoverridecommandlockouts
\usepackage{cite}
\usepackage{amsmath,amssymb,amsfonts}
\usepackage{graphicx}
\usepackage{textcomp}
\usepackage{xcolor}
\usepackage{algorithm}
\usepackage{cite}
\usepackage{amsmath,amssymb,amsfonts}
\usepackage{newunicodechar}
\usepackage{adjustbox}
\usepackage{booktabs}
\usepackage{multirow}
\usepackage{xspace}
\usepackage{amssymb}
\usepackage{algpseudocode}
\usepackage{graphicx}
\usepackage{color,soul}

\newcommand{\Algoname}{ContextGAN\xspace}

\def\BibTeX{{\rm B\kern-.05em{\sc i\kern-.025em b}\kern-.08em
    T\kern-.1667em\lower.7ex\hbox{E}\kern-.125emX}}
\begin{document}

\title{Differentially Private Synthetic Data Generation Using Context-Aware GANs}

\author{\IEEEauthorblockN{Anantaa Kotal}
\IEEEauthorblockA{\textit{Computer Science} \\
\textit{The University of Texas at El Paso}\\
El Paso, US \\
akotal@utep.edu}
\and
\IEEEauthorblockN{Anupam Joshi}
\IEEEauthorblockA{\textit{C.S.E.E.} \\
\textit{University of Maryland, Baltimore County}\\
Baltimore, US \\
joshi@umbc.edu}
}

\maketitle

\begin{abstract}
The widespread use of big data across various sectors has brought significant privacy concerns, particularly when sensitive information is shared or analyzed. Regulations like GDPR and HIPAA impose strict controls on handling  data, making it difficult to balance the need for insights with privacy requirements. Synthetic data offers a promising solution, enabling the creation of artificial datasets that mirror real-world patterns without exposing sensitive information. For instance, synthetic data can simulate patient records or network flows for training machine learning models to conduct research without violating privacy laws. However, traditional synthetic data generation methods often fail to capture complex, implicit rules that relate different elements of the data and are essential in specific domains like healthcare. While these methods might replicate explicit patterns from the training data, they often overlook domain-specific rules that are not directly stated but are critical for maintaining realism and utility. For example, prescription guidelines, such as avoiding certain medications for patients with specific conditions or preventing harmful drug interactions, may not be explicitly represented in the original data. Synthetic data generated without accounting for these implicit rules can lead to medically inappropriate or unrealistic patient profiles. To address these limitations, we propose a framework called Context-Aware Differentially Private Generative Adversarial Network (\Algoname). Our framework integrates domain-specific rules using a constraint matrix that explicitly encodes both explicit and implicit domain knowledge. The constraint-aware discriminator evaluates synthetic data against these rules, ensuring the generated data adheres to domain constraints. Furthermore, the discriminator is differentially private, ensuring privacy preservation by protecting sensitive details from the original data. We validate \Algoname across multiple domains, including healthcare, security, and finance, demonstrating that it produces high-quality synthetic data that respects domain-specific rules while preserving privacy. Our results show that \Algoname significantly improves the realism and utility of synthetic data by enforcing domain constraints, making it suitable for use in scenarios requiring both compliance with explicit patterns and implicit rules, all under strict privacy guarantees.

\end{abstract}

\begin{IEEEkeywords}
Differential Privacy, Synthetic Data, Context Aware Training, GAN
\end{IEEEkeywords}

\section{Introduction}
In the era of big data, the ability to extract meaningful insights from vast and complex datasets has revolutionized fields as diverse as healthcare, security, finance, transportation, and marketing. Machine Learning (ML) models have become essential tools for analyzing these large-scale datasets, enabling advancements in predictive modeling, decision-making, and automated systems. However, the utility of big data is often constrained by privacy and confidentiality concerns. Sharing sensitive data, such as medical records or financial transactions, raises significant privacy and security challenges, particularly when such data must cross organizational or national boundaries. Legal frameworks, including the General Data Protection Regulation (GDPR) and the Health Insurance Portability and Accountability Act (HIPAA), impose strict requirements to protect personal information, forcing organizations to carefully balance data availability and privacy protection \cite{voigt2017eu, gharib2016privacy}.

In healthcare, for example, big data analytics holds enormous potential for improving patient outcomes through data-driven diagnostics, treatment recommendations, and research. However, patient data is highly sensitive, and organizations are often reluctant to share it due to privacy concerns. Privacy-preserving methods are crucial to allow for safe collaboration across institutions. Synthetic data generation has emerged as a promising solution to this dilemma, enabling organizations to share and analyze data without exposing real, sensitive information. Synthetic data replicates the statistical properties of real-world datasets while removing identifying information, thereby reducing privacy risks \cite{goncalves2020generation, abay2019privacy}.

Despite its promise, traditional synthetic data generation methods face limitations, especially in the context of big data. One key issue is that these methods often fail to capture the complex, implicit rules that govern specific domains. In healthcare, for instance, prescription guidelines or treatment pathways may depend on a range of factors, such as age, medical history, and drug interactions, which may not be explicitly encoded in the training data. Without accounting for these implicit relationships, generated data can become unrealistic and may produce misleading results in downstream tasks, such as medical research or clinical trials \cite{choi2017generating}.

Moreover, big data itself presents challenges in terms of volume, variety, and velocity \cite{gandomi2015beyond}. Generative models like Generative Adversarial Networks (GANs) have shown promise in generating synthetic data that closely resembles real-world datasets, but they struggle to scale effectively in domains where domain-specific context plays a critical role \cite{goodfellow2014generative}. In healthcare, for example, models might fail to respect the implicit medical guidelines, such as not prescribing certain medications to patients with specific comorbidities. A synthetic dataset that does not respect these rules could lead to inaccurate predictions or analyses \cite{chen2019learning}. 

To address these limitations, we propose a novel framework called \Algoname (Context-Aware Differentially Private GAN), which leverages the capabilities of Generative Adversarial Networks (GANs) while incorporating domain-specific contextual rules through a constraint matrix. \Algoname goes beyond traditional synthetic data generation methods by ensuring that the generated data not only mirrors the explicit statistical properties of the original dataset but also respects implicit domain rules critical for maintaining realism and utility. For instance, in the healthcare domain, \Algoname can generate synthetic patient records that adhere to medical guidelines for drug prescriptions, even when these guidelines are not explicitly present in the training data.

Our framework includes a constraint matrix, which encodes both explicit and implicit domain knowledge, guiding the GAN's discriminator to identify and penalize data that violates these rules. This constraint-aware discriminator ensures that generated data aligns with critical domain-specific rules. Additionally, \Algoname integrates differential privacy techniques to protect individual-level information during training. The discriminator is designed to be differentially private, adding noise to the training process and ensuring that no sensitive information from the original dataset is leaked into the generated synthetic data.\Algoname addresses the limitations of standard GANs by incorporating domain-specific context, making it particularly effective in fields like healthcare, finance, and security, where implicit rules are essential for maintaining data fidelity. The combination of differential privacy and context-aware mechanisms ensures that the generated synthetic data can be safely shared and analyzed without compromising either privacy or domain-specific realism.

We validate our framework through extensive experiments across multiple domains, showing that \Algoname consistently produces high-quality synthetic data that is both privacy-preserving and contextually accurate. By balancing differential privacy with domain-specific context, \Algoname ensures that the synthetic data can be safely utilized in big data environments, offering significant utility for organizations bound by stringent privacy constraints. This makes \Algoname a valuable tool for generating realistic, safe, and useful synthetic data in scenarios requiring both compliance with explicit patterns and implicit domain-specific rules.

The remainder of this paper is structured as follows: In Section \ref{sec_background}, we review the background and motivation for synthetic data generation in the context of big data and privacy. Section \ref{sec_proposed_framework} outlines the \Algoname framework and its technical foundations. In Section \ref{sec_exp_results}, we demonstrate \Algoname's effectiveness in preserving both privacy and utility across various big data applications. Finally, Section \ref{sec_conclusion} offers concluding remarks and discusses future directions for research.

\section{Background}
\label{sec_background}
\subsection{Data Synthesis for Privacy}
In addressing data privacy challenges in distributed systems, several strategies have been developed, including data anonymization, secure cryptographic methods, and distributed model release. Each approach, however, has limitations. Data anonymization, though widely used, can be vulnerable to re-identification attacks when new information is combined with the anonymized dataset \cite{sweeney2002k, machanavajjhala2007diversity, dwork2006calibrating, rocher2019estimating}. Cryptographic methods like Secure Multiparty Computation ensure secure data sharing but restrict open access, limiting usability. Distributed model release, while effective in specific scenarios, often incurs high costs and centralization issues, hindering seamless data sharing across diverse organizations.

Synthetic data generation offers a promising alternative by creating new datasets that maintain statistical similarities to real data but lack direct individual associations. This minimizes the risk of privacy breaches while allowing organizations to share data for research or collaboration \cite{goncalves2020generation, abay2019privacy}. In fields like healthcare, this approach is particularly valuable for ensuring that sensitive micro-data can be shared without exposing individual-level information.

Generative models used in synthetic data generation aim to replicate the statistical properties of real datasets, making them indistinguishable from original data for analytical tasks. This encourages data sharing, reduces privacy risks, fosters experimentation, and accelerates project timelines.

Differential privacy, widely regarded for its simplicity and efficiency, is the most common framework for protecting privacy in machine learning models \cite{dwork2006calibrating}. Differential privacy ensures that the output of an algorithm remains largely unchanged even if one data point is altered.  Differential Privacy Stochastic Gradient Descent (DP-SGD) \cite{abadi2016deep}, a privacy-preserving deep learning algorithm, adds noise to gradient computations during each iteration, protecting sensitive information from inference attacks.

Other advanced methods include the Probability Graphical Models framework \cite{mckenna2019graphical}, which employs a blend of parametric and non-parametric techniques to estimate parameters, and PrivBayes \cite{zhang2017privbayes}, which uses Bayesian networks to model data distribution. Abay et al. \cite{abay2019privacy} propose a privacy-preserving autoencoder-based synthesis method, injecting noise into latent representations to protect privacy during data generation.

\subsection{Privacy-Preserving Data Generation using GANs}
Generative adversarial Networks (GANs) is a Generative Deep Learning model that have been successfully used to generate synthetic data that bear strong statistical similarities to the original data. GANs have been shown to be extremely accurate in synthetic data generation and translation, particularly for image and text data \cite{brock2018large, isola2017image, zhang2017stackgan, wang2018high}. The principal architecture in a GAN framework involves a generative model G that captures the data distribution, and a discriminative model D that estimates the probability that a sample came from the original distribution rather than G. The training procedure for G is to maximize the probability of D making a mistake. As the training progresses, the generator gets better at generating new examples that plausibly come close to the samples from the original distribution. The idea behind GAN can be formulated as a two-player min-max game with value function $V (G, D)$:
\begin{equation}
    \begin{aligned}
        {\underset{G}{\min}}\,{\underset{D}{\max}}\, V(G,D)= E_{x \sim p_{data}(x)}[log D(x)] 
        \\ + E_{z \sim p_z(z)}[log(1 - D(G(z)))]
    \end{aligned}
\end{equation}

Teaching Generative Adversarial Networks (GANs) to learn from Network Activity Data poses challenges due to its inherent characteristics. Typically tabular, this data consists of a combination of discrete and continuous values. Moreover, it frequently exhibits sparsity, with attribute values unevenly distributed, leading to imbalances in the dataset. These complexities in structure and distribution make it a demanding task for GANs to effectively capture and reproduce the patterns present in Network Activity Data.

In the 2019 paper by Xu et al. \cite{xu2019modeling}, the author proposes a GAN model to address the challenges with tabular data. Specifically, they address the issues of multi-modality and non-gaussian nature of continuous variables and class imbalance in tabular data. The authors propose 3 key steps to handle these challenges: (1) Mode-specific normalization, (2) Conditional Generator, and (3) Training by sampling. In the 2021 paper by Kotal et al.\cite{kotal2022privetab}, further used this model to create a framework for privacy preserving data generation in tabular data. The authors enforce t-closeness in the generated data to the original dataset to preserve privacy. For this, the Earth Mover's distance (EMD) of the distribution of features in the synthetic is calculated w.r.t. the original dataset. 

Differential Privacy, like other deep learning models, can also be applied to the GAN framework to enhance privacy. Building on Abadi et al.'s work \cite{abadi2016deep}, various GAN models have been proposed that combine Differential Privacy Stochastic Gradient Descent (DP-SGD) with Generative Adversarial Networks (GANs) to generate differentially private synthetic data \cite{xie2018differentially, torkzadehmahani2019dp}. These models introduce noise into the GAN's discriminator during the training process to enforce differential privacy. The crucial aspect of DP's guarantee of post-processing privacy means that by safeguarding the GAN's discriminator, it ensures differential privacy for the parameters of the GAN's generator. 

Our previous work has made significant strides in privacy-preserving synthetic data generation \cite{kotal2023privacy,garza2024privcomp,das2023change}. The Privetab framework \cite{kotal2022privetab} utilized a Conditional GAN to generate synthetic data while ensuring t-closeness for privacy. KiNETGAN \cite{kotal2024kinetgan} extended this by introducing a knowledge-guided discriminator, enforcing domain-specific constraints for network intrusion detection. Additionally, the KIPPS framework \cite{kotal2024kipps} integrated predefined domain rules into the generator loss, ensuring rule adherence during synthetic data generation.

Building on these foundations, our new model, ContextGAN, brings two key advancements. First, it leverages a constraint matrix to explicitly encode domain-specific rules and integrate them directly into the discriminator's loss function, allowing for more precise rule enforcement during training. Second, unlike our previous work, ContextGAN ensures that the discriminator is differentially private, offering enhanced privacy protection while maintaining compliance with domain rules, resulting in more accurate and privacy-preserving synthetic data.

\subsection{Knowledge Guided Learning} 
GANs relying solely on observed network activity data possess a restricted understanding of network attributes such as IP addresses and domain URLs. In the network activity domain, strict rules govern actions, such as the legitimate range of port numbers associated with a protocol. GANs cannot inherently adhere to these rules without explicit constraints. Knowledge guidance proves essential in explicitly conveying these constraints to the generative model, enabling it to generate data that aligns with the established rules and ensuring the realism and adherence to protocol necessary for meaningful and accurate synthetic data generation within the observed system.

Knowledge Graphs (KG) serve as a versatile graph-structured data model designed for knowledge representation and reasoning. Within a KG, information is organized into semantic triples, consisting of a subject, predicate, and object. Subjects and objects correspond to nodes or entities in the graph, while predicates denote the functional relationships between them. The graph-like structure of KGs allows for the efficient storage of extensive information, and the network can be expanded with new knowledge as required. KGs are equipped with powerful reasoning capabilities, enabling the imposition of constraints on entities and the deduction of new knowledge. For instance, specifying constraints such as the source IPs must not belong to a subnet or must originate from a specific external CIDR range. Leveraging these robust knowledge representation and reasoning abilities, KGs can enhance and enrich the data generation process.

Knowledge graphs excel at storing contextual information crucial for enhancing learning in distributed systems. The Unified Cybersecurity Ontology (UCO) stands out as a comprehensive ontology designed for cyber situational awareness in cybersecurity systems. Its integration has been demonstrated to significantly improve contextual awareness in machine learning (ML) systems, as evidenced by studies such as those by Piplai et al. \cite{piplai2020creating} and Narayanan et al. \cite{narayanan2018early}. In a related context, Hui et al. \cite{hui2022knowledge} have introduced a knowledge-enhanced Generative Adversarial Network (GAN) designed to generate Internet of Things (IoT) traffic data for devices from various manufacturers. Their approach however is specific to IoT traffic and uses knowledge injection to set the conditions for IoT traffic generation. In this paper, we inject knowledge about network traffic into the training of the GAN by adding the Knowledge base as an independent discriminator.

\section{Proposed Framework}
\label{sec_proposed_framework}
To address the limitations of traditional synthetic data generation methods, we propose the \Algoname\ (Differentially Private Context-Aware GAN) framework, which effectively incorporates domain-specific constraints into the synthetic data generation process. This framework leverages the power of Generative Adversarial Networks (GANs) while ensuring that the generated data reflects not only explicit statistical patterns but also critical implicit rules that are essential for maintaining realism and utility, especially in domains like healthcare, finance, and security.
A central challenge in many domains, particularly healthcare, is that data often contains implicit rules—such as treatment protocols or drug interaction guidelines—that are not always directly encoded in raw datasets. These rules must be enforced in the synthetic data to ensure it is realistic and applicable for downstream tasks such as clinical research, predictive modeling, or policy development. For example, in healthcare, prescription guidelines may prohibit certain drugs from being prescribed together due to harmful interactions. These relationships are often documented in domain knowledge bases but are not always evident from the dataset alone.

\Algoname\ addresses this challenge by integrating domain-specific constraints into the training process via a constraint matrix, which encodes both explicit and implicit rules. This matrix ensures that the synthetic data adheres to real-world domain constraints, thus improving the quality and realism of the generated data. The constraint matrix is incorporated into the GAN’s discriminator, guiding the model to reject synthetic data that violates the encoded rules.

The \Algoname\ model achieves two main objectives through its key innovations:

\begin{enumerate}
   \item \textbf{Enforcing Domain-Specific Rules in Generated Data:} The framework uses a context-aware discriminator guided by a constraint matrix. This matrix encodes domain-specific knowledge, such as medical guidelines or regulatory requirements, that define valid and invalid attribute combinations. For instance, in healthcare, the matrix ensures that generated patient records conform to rules governing which drug combinations are safe based on patient demographics, medical history, or allergies. The discriminator evaluates whether the generated data violates any domain-specific rules by checking the generated data against the constraint matrix. If a generated data point violates these rules (e.g., a contraindicated drug combination), the discriminator penalizes the generator. This process forces the generator to produce data that respects both explicit and implicit domain rules, ensuring that synthetic data is realistic and applicable in practice.
    \item \textbf{Ensuring Privacy with a Differentially Private Discriminator:} To ensure that the synthetic data generation process complies with privacy regulations such as GDPR and HIPAA, the framework integrates differential privacy mechanisms. The discriminator is made differentially private by adding carefully calibrated noise to the gradients during optimization, ensuring that sensitive information from the original dataset is not leaked into the generated data. By combining differential privacy with the constraint matrix, the model protects against privacy attacks, such as re-identification or attribute inference, while still producing high-quality, realistic synthetic data.

\end{enumerate}
    
By incorporating explicit domain rules into the discriminator through the constraint matrix, the framework ensures that the synthetic data conforms to important real-world constraints, making it realistic and useful for applications like predictive modeling or research. The addition of differential privacy further ensures that sensitive data is protected, making the model suitable for generating synthetic datasets in privacy-sensitive domains like healthcare.

\textbf{Development of the Constraint Matrix:} The \textit{constraint matrix} serves as a formal mechanism to encode domain-specific rules, such as medical guidelines or regulatory requirements. Let the dataset \( X \) consist of records with multiple attributes \( \{x_1, x_2, \dots, x_n\} \), where each attribute represents a domain-specific feature (e.g., age, diagnosis, treatment). The constraint matrix \( CM \) encodes whether a specific combination of attribute values is valid or invalid based on domain rules. 

Formally, the constraint matrix \( CM \) is defined as:

\[
CM: \mathcal{X}_1 \times \mathcal{X}_2 \times \dots \times \mathcal{X}_n \rightarrow \{0, 1\}
\]

where \( \mathcal{X}_i \) represents the possible values of the \( i \)-th attribute, and \( CM(x_1, x_2, \dots, x_n) = 1 \) indicates a valid combination, while \( CM(x_1, x_2, \dots, x_n) = 0 \) indicates an invalid combination. This matrix provides a binary output for each attribute combination, enforcing domain-specific rules efficiently.

\textbf{Integrating the Constraint Matrix into the Discriminator:} The constraint matrix is integrated into the discriminator’s loss function to guide the GAN in generating data that adheres to domain-specific rules. For each generated data point \( x_{\text{gen}} = (x_1, x_2, \dots, x_n) \), the discriminator checks the validity of the attribute combination using the constraint matrix:

\[
\text{CM}(x_{\text{gen}}) = CM(x_1, x_2, \dots, x_n)
\]

This check returns 1 if the data point satisfies domain constraints and 0 otherwise. The discriminator penalizes invalid combinations using a modified loss function that incorporates the constraint matrix. The loss function for the discriminator is given by:

\[
L_D = \mathbb{E}_{x_{\text{real}}}[\log(D(x_{\text{real}}))] + \mathbb{E}_{x_{\text{gen}}}[\log(1 - D(x_{\text{gen}}))] - \lambda \cdot \text{CM}(x_{\text{gen}})
\]

where \( \lambda \) is a hyperparameter controlling the penalty for violating domain rules, \( D(x) \) represents the discriminator’s output for data point \( x \), and \( \text{CM}(x_{\text{gen}}) \) is the result of the constraint matrix evaluation for the generated data point. This formulation ensures that the generator is guided to produce synthetic data that is both realistic and compliant with domain rules.

As training progresses, the discriminator evaluates both real and generated data points. The discriminator looks up each generated data point in the constraint matrix and determines whether it satisfies the domain-specific rules. If the generated data violates any of the encoded rules, the discriminator penalizes it in proportion to the violation, which forces the generator to adjust its output in subsequent iterations to comply with the domain constraints.

\textbf{Making the Discriminator Differentially Private:} To ensure privacy preservation, we adopt the \textit{Differential Privacy Stochastic Gradient Descent (DP-SGD)} approach as proposed by \cite{abadi2016deep}. The goal is to guarantee that individual data points from the original dataset cannot be inferred during the training process. Differential privacy in the discriminator is achieved through two primary mechanisms: gradient clipping and noise addition.

During each training iteration, the gradients of the discriminator’s loss function are first clipped to a predefined threshold \( C \), which ensures that no single data point has a disproportionate influence on the model’s parameters. Formally, the clipped gradient \( \nabla \theta \hat{L}_D \) is computed as follows:

\[
\nabla \theta \hat{L}_D = \frac{\nabla \theta L_D}{\max\left(1, \frac{||\nabla \theta L_D||_2}{C}\right)}
\]
The differentially private gradient update is given by:

\[
\nabla \theta \hat{L}^{priv}_D = \nabla \theta\hat{L}_D + \mathcal{N}(0, \sigma^2)
\]
where \( \mathcal{N}(0, \sigma^2) \) represents Gaussian noise with mean 0 and variance \( \sigma^2 \), which is calibrated based on the differential privacy budget \( \epsilon \). By adding noise to the clipped gradients, we ensure that the discriminator remains differentially private, which prevents sensitive information from being leaked during training. This mechanism guarantees that the inclusion or exclusion of a single data point does not significantly affect the output of the discriminator, thereby protecting the privacy of individuals in the dataset.

\Algoname\ effectively balances privacy preservation with domain-specific accuracy by integrating a constraint matrix into the GAN’s discriminator and applying differential privacy. The framework is designed to generate synthetic data that adheres to essential real-world rules while ensuring that individual data privacy is maintained. This approach allows \Algoname\ to produce realistic and useful synthetic datasets that can be applied across various sensitive domains such as healthcare, finance, and security. The complete algorithm for the \Algoname\ model, including its differential privacy and constraint-aware components, can be found in \ref{alg:ContextGAN}.

\begin{algorithm}[ht]
\caption{\Algoname: Differentially Private Context-Aware GAN with Constraint Matrix}
\label{alg:ContextGAN}
\textbf{Input:} Real dataset $X = \{x_1, x_2, \dots, x_m\}$, constraint matrix $CM$, privacy parameters $(\epsilon, \delta)$, gradient clipping threshold $C$, noise variance $\sigma^2$, learning rate $\eta$\\
\textbf{Output:} Generator $G$ that produces synthetic data adhering to domain constraints

\begin{algorithmic}[1]

\State \textbf{Initialize} generator $G$ and discriminator $D$ with random parameters $\theta_G$ and $\theta_D$.

\For {each iteration}
    \State Sample a minibatch of real data points $\{x_1, x_2, \dots, x_n\}$ from $X$.
    \State Sample a minibatch of random noise $\{z_1, z_2, \dots, z_n\}$ from a noise distribution (e.g., uniform or Gaussian).
    \State Generate a minibatch of synthetic data $\{x_{\text{gen},1}, x_{\text{gen},2}, \dots, x_{\text{gen},n}\}$ using the generator $G(z_i)$.
    
    \State \textbf{Discriminator update:}
    \For {each generated data point $x_{\text{gen}} = (x_1, x_2, \dots, x_n)$}
        \State Look up constraint matrix $CM(x_{\text{gen}})$ to check if the generated data is valid.
        \State Compute discriminator loss:
        \[
        L_D = \mathbb{E}_{x_{\text{real}}}[\log(D(x_{\text{real}}))] + \mathbb{E}_{x_{\text{gen}}}[\log(1 - D(x_{\text{gen}}))] - \lambda \cdot \text{CM}(x_{\text{gen}})
        \]
        \State Clip gradients of discriminator's loss:
        \[
        \nabla \theta_D \hat{L}_D = \frac{\nabla \theta_D L_D}{\max\left(1, \frac{||\nabla \theta_D L_D||_2}{C}\right)}
        \]
        \State Add noise to the clipped gradients to ensure differential privacy:
        \[
        \nabla \theta_D^{\text{priv}} = \nabla \theta_D \hat{L}_D + \mathcal{N}(0, \sigma^2)
        \]
        \State Update discriminator parameters:
        \[
        \theta_D \gets \theta_D - \eta \nabla \theta_D^{\text{priv}}
        \]
    \EndFor
    
    \State \textbf{Generator update:}
    \State Compute generator loss (based on discriminator feedback):
    \[
    L_G = \mathbb{E}_{z_i}[\log(1 - D(G(z_i)))]
    \]
    \State Update generator parameters using the gradient of the generator loss:
    \[
    \theta_G \gets \theta_G - \eta \nabla \theta_G L_G
    \]
    
\EndFor

\State \textbf{Return} Generator $G$.

\end{algorithmic}
\end{algorithm}

\section{Evaluation}
\label{sec_exp_results}
In this section, we detail the evaluation of the \Algoname\ framework across datasets from the healthcare, security, and finance domains. These datasets support machine learning-based classifiers, which require high-quality data for training. Obtaining reliable and representative training data is often a significant challenge in these domains, so synthetic data must effectively substitute the original data in downstream tasks such as classification and prediction. To demonstrate that the \Algoname\ framework achieves these goals, we performed evaluations across three key metrics:

\begin{itemize}
    \item \textbf{Fidelity Results}: Demonstrating that the synthetic data is statistically close to the original data.
    \item \textbf{Utility Results}: Evaluating whether the synthetic data is useful in training downstream machine learning classifiers in healthcare, security, and finance domains.
    \item \textbf{Privacy Results}: Assessing the resilience of the synthetic data and the \Algoname\ model against privacy attacks, such as re-identification, attribute inference, and membership inference.
\end{itemize}

We evaluated the performance of \Algoname\ against other state-of-the-art generative models for tabular data, including CTGAN \cite{xu2019modeling}, TVAE \cite{xu2019modeling}, PATEGAN \cite{jordon2018pate}, and TABLEGAN \cite{park2018data}, across six datasets from the healthcare, security, and finance domains.

\subsection{Datasets}
To evaluate the effectiveness of the \Algoname\ framework, we conducted experiments using six publicly available datasets from three critical domains: healthcare, security, and finance. These datasets were chosen for their relevance in supporting machine learning tasks that require high-quality and privacy-sensitive data.

\begin{itemize}
    \item \textbf{PIMA Indians Diabetes Dataset}: This dataset contains 768 records from the Pima Indian population, with 8 features related to medical measurements such as age, blood pressure, and glucose levels. The task is to predict the likelihood of diabetes onset based on diagnostic measurements.
    \item \textbf{Heart Disease UCI Dataset}: This dataset consists of 303 records, each with 14 attributes including patient demographics and medical data. The objective is to predict the presence of heart disease.
    \item \textbf{UNSW-NB15 Dataset}: A benchmark dataset for network intrusion detection, containing network traffic data with 49 features that describe different attributes of the connections. The reduced version of the dataset contains over 175,000 records.
    \item \textbf{CICIDS2017 Dataset}: A widely used dataset for evaluating intrusion detection systems, which captures various types of malicious activities in network traffic. The reduced version includes a subset of the full dataset, focusing on specific attack types and network behaviors.
    \item \textbf{UCI Credit Card Default Dataset}: This dataset contains 30,000 records from a Taiwanese bank, each with 24 features, including demographic information, credit history, and bill payment records. The task is to predict whether a customer will default on their credit card payment for the following month.
    \item \textbf{Statlog (German Credit) Dataset}: This dataset contains 1,000 records with 20 attributes that represent financial and personal data. The task is to classify individuals as either having good or bad credit risk.
\end{itemize}

These datasets encompass a range of classification tasks across different domains, ensuring that the \Algoname\ framework was tested on diverse data structures, including both continuous and categorical variables. The experiments on these datasets allowed us to evaluate the performance of \Algoname\ in terms of fidelity, utility for machine learning models, and privacy preservation, providing a comprehensive assessment of its applicability in real-world scenarios.

\begin{table}[!htb]
\centering
\resizebox{0.8\columnwidth}{!}{%
\begin{tabular}{|l|l|l|}
\hline
                                  & \textbf{EMD} & \textbf{Distance} \\ \hline
\textbf{\Algoname} & 0.12         & 0.15              \\ \hline
\textbf{CTGAN}                    & 0.14         & 0.15              \\ \hline
\textbf{TVAE}                     & 0.13         & 0.16              \\ \hline
\textbf{PateGAN}                  & 1.25         & 1.10              \\ \hline
\textbf{TableGAN}                 & 1.40         & 1.25              \\ \hline
\end{tabular}%
}
\caption{Comparison of Distance between Synthetic and Original Data}
\label{tab:distance}
\end{table}

\subsection{Fidelity Results}
To assess the quality of the generated synthetic data, we calculated the statistical distance between the original and generated data distributions. Specifically, we used two distance metrics:

\begin{itemize}
    \item \textbf{Earth Mover's Distance (EMD)} or Wasserstein Distance: This metric measures the minimum cost of transforming one distribution into another and is particularly useful for continuous data distributions.
    \item \textbf{Combination of $L_1$ and $L_2$ Norms}: For tabular data containing both categorical and continuous variables, we used the $L_1$ norm (Manhattan distance) for categorical variables and the $L_2$ norm (Euclidean distance) for continuous variables. This approach is well-suited for mixed-type datasets.
\end{itemize}

The average of the distances for all 6 datasets have been summarized in Table \ref{tab:distance}. The results demonstrate that \Algoname\ consistently outperformed or matched the performance of other models, such as CTGAN and TVAE, across all datasets. \Algoname\ exhibited the lowest Earth Mover's Distance (EMD) and combined distance in the healthcare, security, and finance datasets, indicating that the generated synthetic data closely resembles the original data.

\begin{figure*}
    \centering
    \includegraphics[width=0.7\textwidth]
    {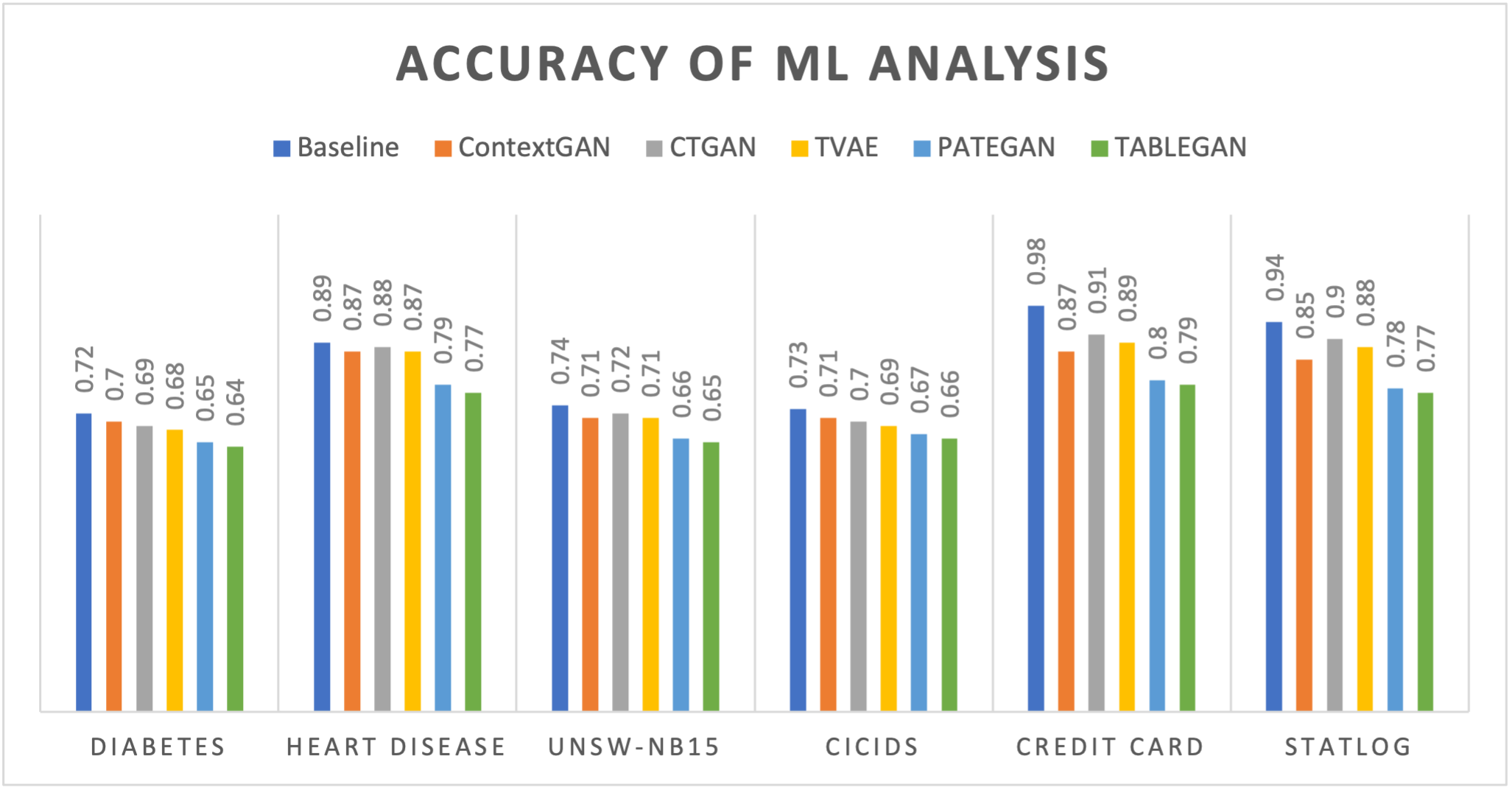}
    \caption{Comparison of NIDS accuracy for Lab Collected Data}
    \label{fig:accuracy}
\end{figure*}

\subsection{Utility Results}
To test the utility of the synthetic data, we evaluated its performance in training machine learning classifiers for tasks in healthcare, security, and finance. For each dataset, we trained standard classification models such as Random Forest, XGBoost, and Logistic Regression on the synthetic data generated by \Algoname\ and compared the accuracy, precision, recall, and F1 scores with models trained on the original data. We also compared these results with models trained on synthetic data generated by CTGAN, OCTGAN, PATEGAN, and TABLEGAN.

Figure \ref{fig:accuracy} demonstrates that \Algoname\ consistently achieves high accuracy across various datasets, particularly in challenging domains like healthcare, security, and finance. Compared to the baseline model, \Algoname\ performs competitively, maintaining accuracy levels close to the baseline, especially in Diabetes, UNSW-NB15, and CICIDS datasets.

In comparison to other generative models such as CTGAN and TVAE, \Algoname\ consistently delivers comparable or slightly better performance, reinforcing its robustness in generating high-quality synthetic data that retains significant utility for downstream tasks. Models like PATEGAN and TABLEGAN tend to perform lower across all datasets, further highlighting the efficiency of \Algoname\ in balancing domain-specific constraints and privacy while ensuring reliable performance. 

\subsection{Privacy Results}

To assess the privacy-preserving properties of \Algoname\ and the generated synthetic data, we evaluated the model's resilience against several common privacy attacks, including re-identification, attribute inference, and membership inference attacks.

\begin{figure}[ht]
    \centering
    \includegraphics[width=\columnwidth]
    {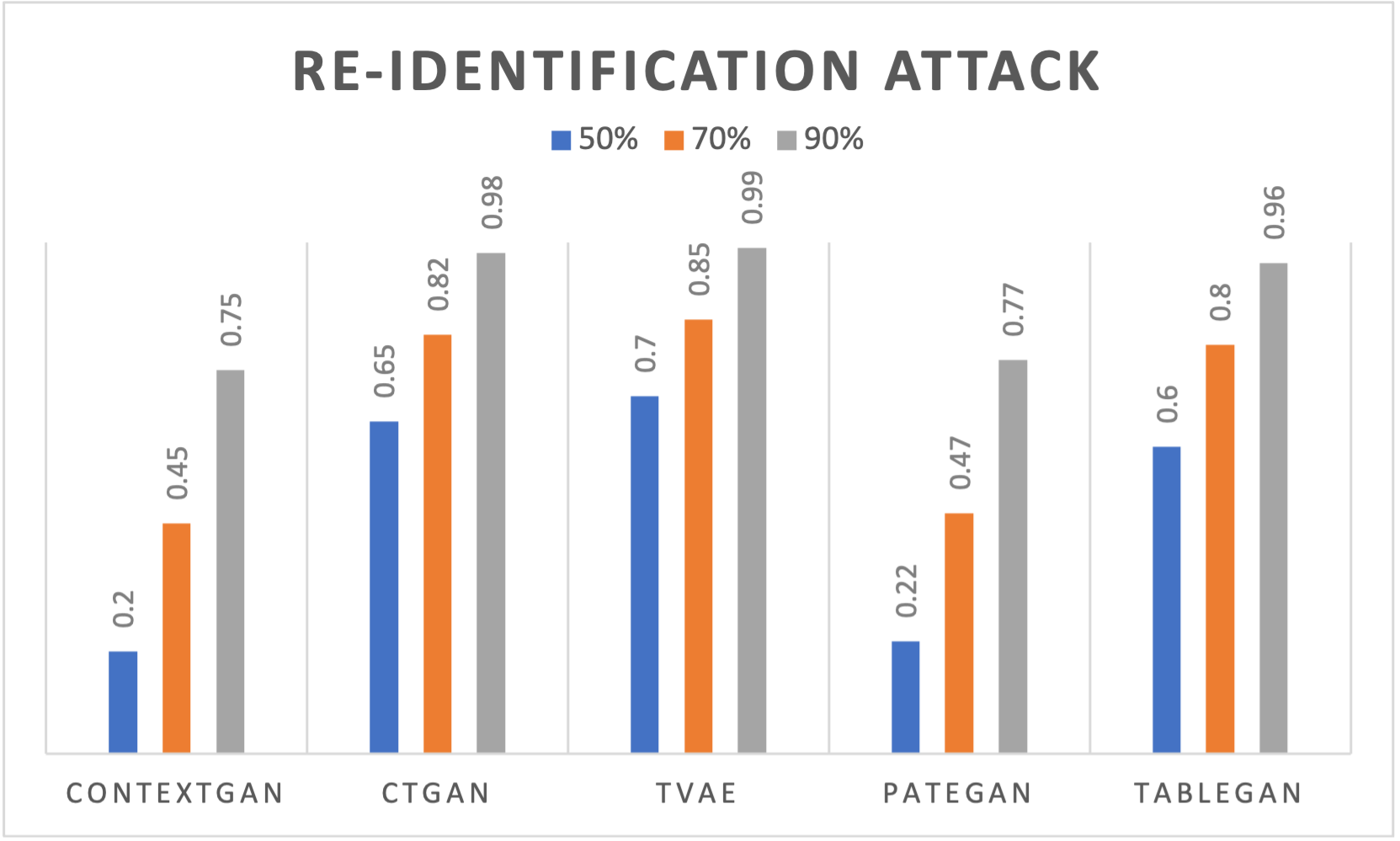}
    \caption{Comparison of Re-identification Attack with 30\%, 60\% and 90\% overlap on original data}
    \label{fig:reident}
\end{figure}

\textbf{Re-identification Attacks:} Re-identification attacks attempt to link de-identified synthetic data back to real-world identities using prior knowledge. We evaluated the attack success rate by varying the overlap between the adversary's knowledge and the dataset. In our experiment, we designed the re-identification attack by comparing the attribute values of the synthetic dataset to the attribute values of the original dataset in order to locate potential examples from the original dataset that have leaked into the synthetic dataset. We assume that the attacker has partial knowledge of the dataset or has access to external data with partial overlap with the sensitive dataset. With this partial knowledge, we try to find matches in the generated data. As shown in Figure \ref{fig:reident}, \Algoname\ and PATEGAN consistently demonstrated the lowest attack success rates across all overlap levels, indicating strong mitigation of re-identification risks. At 90\% overlap, \Algoname\ achieved a success rate of 0.75, compared to 0.98 for CTGAN and 0.99 for TVAE. Similarly, at 70\% overlap, \Algoname\ had a success rate of 0.45, significantly lower than CTGAN (0.82) and TVAE (0.85), further confirming \Algoname's effectiveness in preserving privacy.

\begin{figure}[ht]
    \centering
    \includegraphics[width=\columnwidth]
    {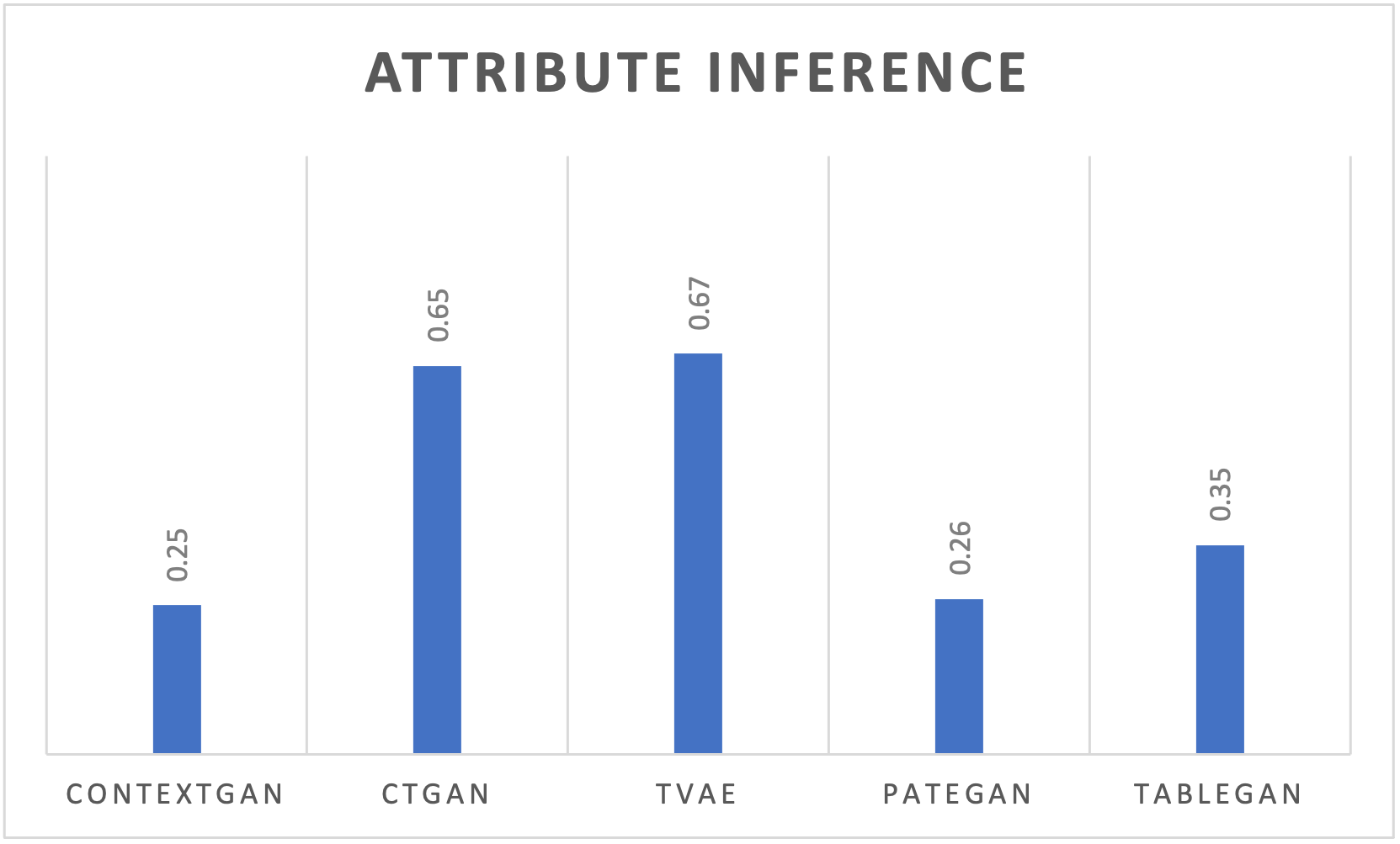}
    \caption{Comparison of accuracy in Attribute Inference Attack}
    \label{fig:attrib_infer}
\end{figure}

\textbf{Attribute Inference Attacks:} Attribute inference attacks seek to predict sensitive attributes (e.g., health conditions or financial status) from other, non-sensitive data points. In our experiment, we follow the black box model inversion attribute inference attack as described by Mehnaz et al. \cite{mehnaz2020black}. The attack follows a Confidence modeling based Model inversion attack, where the adversary randomly predicts the sensitive attribute by assigning confidence scores to each possible value.  Figure \ref{fig:attrib_infer} shows that \Algoname\ demonstrated strong resilience against attribute inference attacks, with an attack accuracy of only 0.25, significantly lower than other models like TABLEGAN (0.35) and PATEGAN (0.26). In contrast, models like CTGAN and TVAE exhibited much higher attack success rates, with accuracies of 0.65 and 0.67, respectively, highlighting \Algoname's superior ability to protect sensitive information.

\begin{figure}[ht]
    \centering
    \includegraphics[width=\columnwidth]
    {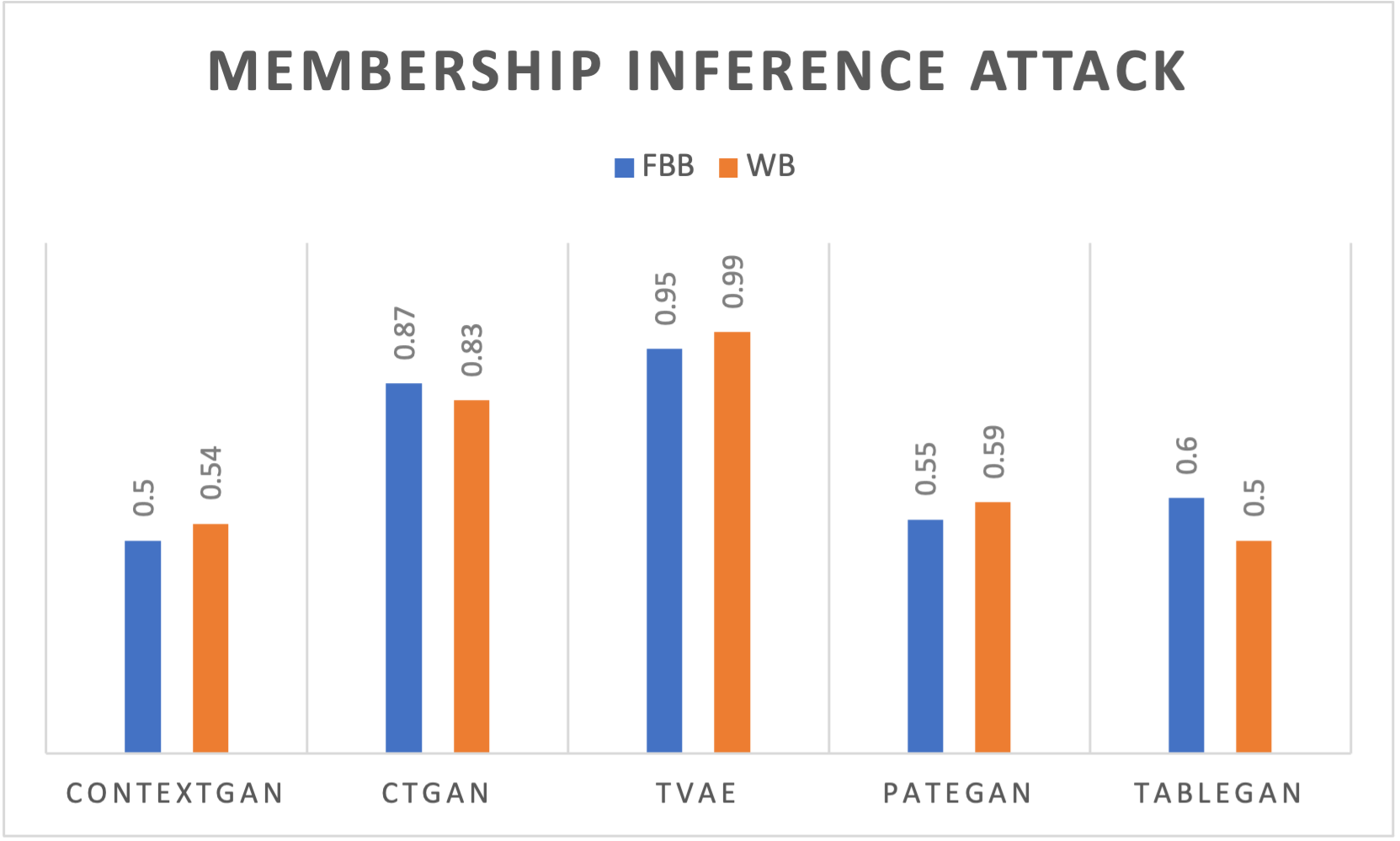}
    \caption{Comparison of Membership Inference Attack in White Box (WB) and FBB (Fully Black Box) setting}
    \label{fig:mia}
\end{figure}

\textbf{Membership Inference Attacks:} A Membership Inference attack is a privacy threat specifically targeting machine learning models. The attack aims to determine whether a specific data point was part of the training dataset used to build the model. The adversary, with access to the model's predictions, exploits the model's output to infer the presence or absence of an individual data point in the training set.
The attack unfolds in two distinct phases. In the Training Phase, the adversary systematically collects intelligence about the target model by submitting diverse inputs. Concurrently, the adversary crafts a shadow dataset, mirroring the characteristics of the original training dataset, although not necessarily identical. The subsequent Inference Phase leverages the information amassed during training. Here, the adversary endeavors to determine whether a particular data point was part of the model's original training set. 
In our experiment, we follow the Membership Inference attack framework as described by Chen et al. \cite{chen2020gan}. We assume two settings for the adversary. In a White-Box (WB) setting, the adversary has comprehensive knowledge about the architecture, parameters, and training data of the target model. This means the attacker essentially has access to the internal workings of the model. In a Fully Black Box (FBB) setting, the adversary has limited or no information about the inner workings of the target model. They only have access to the model's input-output behavior without knowledge of its architecture, parameters, or training data. We evaluated \Algoname\ (ContextGAN) in both white-box (WB) and fully black-box (FBB) settings. As shown in Figure \ref{fig:mia}, \Algoname\ demonstrated strong robustness in both scenarios, achieving an attack accuracy of 0.50 in the FBB setting and 0.54 in the WB setting, significantly outperforming models like CTGAN and TVAE, which had much higher attack success rates (CTGAN: 0.87 FBB, 0.83 WB; TVAE: 0.95 FBB, 0.99 WB). Compared to other models like PATEGAN and TABLEGAN, \Algoname\ showed consistently lower attack success rates, confirming its superior privacy preservation.

\section{Conclusion and Future work}
\label{sec_conclusion}
In this paper, we introduced \Algoname, a novel framework for generating privacy-preserving synthetic data that integrates a constraint matrix to explicitly enforce domain-specific rules while ensuring differential privacy in the discriminator. This approach guarantees that the generated data adheres to critical domain constraints while protecting sensitive information from privacy attacks. We evaluated \Algoname across six datasets from the healthcare, security, and finance domains, demonstrating its effectiveness in generating realistic data that closely mirrors the original datasets and performs well in downstream machine learning tasks. \Algoname consistently delivered competitive or superior performance compared to state-of-the-art models. Moreover, \Algoname exhibited strong resilience against privacy attacks, including re-identification, attribute inference, and membership inference, proving its capability to safeguard privacy while generating useful data. 

Its applicability across multiple domains highlights its flexibility and effectiveness in addressing the challenges of data scarcity and privacy concerns, making it particularly suitable for privacy-sensitive fields like healthcare, finance, and security. For future work, we aim to explore further refinement of the constraint matrix, particularly in more complex domains where rule interdependencies may play a crucial role. Additionally, expanding the differential privacy mechanisms to provide more granular control over privacy budgets and testing the framework in more diverse datasets and contexts will further enhance \Algoname's applicability and robustness. These advancements will continue to improve the framework’s capability to generate high-quality, privacy-preserving synthetic data tailored to specific domain needs.

To further enhance the comprehensiveness of our evaluation, future work could incorporate additional, recent models that have introduced notable advancements in synthetic data generation. For instance, SynthPop\cite{nowok2016synthpop}, MST \cite{mckenna2021winning}, and the Avatar models \cite{guillaudeux2023patient, lebrun2024synthetic} represent valuable contributions with unique approaches towards privacy preserving synthetic data generation. Including these models in future comparisons would provide deeper insights and allow for a broader evaluation of synthetic data methodologies. Additionally, these models will be discussed in an expanded literature review, and their potential integration into experimental designs will be explored within the scope of subsequent research.

\bibliographystyle{IEEEtran}
\bibliography{ref}

\end{document}